\documentclass[extendedabs]{recpad2k}

\title{Zero-Shot Action Recognition in Surveillance Videos}

\addauthor{João Pereira}{jaca.pereira@campus.fct.unl.pt}{1}
\addauthor{Vasco Lopes}{vasco.lopes@ubi.pt}{3}
\addauthor{David Semedo}{df.semedo@fct.unl.pt}{1}
\addauthor{João Neves}{jcneves@di.ubi.pt }{2}

\addinstitution{
 NOVALINCS, NOVA School of Science and Technology \\
2829-516 Caparica, PT
}

\addinstitution{
 NOVALINCS, Universidade da Beira Interior \\
 Rua Marquês D’Ávila e Bolama \\
    6200-001, Covilhã, PT
}

\addinstitution{
DeepNeuronic \\
 Covilhã, PT
}

\runninghead{Student, Prof, Collaborator}{RECPAD Author Guidelines}

\usepackage{booktabs}
\usepackage{multirow}
\usepackage{amsmath}
\usepackage{amssymb}

\begin{document}

\maketitle

\begin{abstract}
The growing demand for surveillance in public spaces presents significant challenges due to the shortage of human resources. Current AI-based video surveillance systems heavily rely on core computer vision models that require extensive finetuning, which is particularly difficult in surveillance settings due to limited datasets and difficult setting (viewpoint, low quality, etc.). In this work, we propose leveraging Large Vision-Language Models (LVLMs), known for their strong zero and few-shot generalization, to tackle video understanding tasks in surveillance. Specifically, we explore VideoLLaMA2, a state-of-the-art LVLM, and an improved token-level sampling method, Self-Reflective Sampling (Self-ReS). Our experiments on the UCF-Crime dataset show that VideoLLaMA2 represents a significant leap in zero-shot performance, with 20\% boost over the baseline. Self-ReS additionally increases zero-shot action recognition performance to 44.6\%. These results highlight the potential of LVLMs, paired with improved sampling techniques, for advancing surveillance video analysis in diverse scenarios.
\end{abstract}

\section{Introduction}
\label{sec:intro}
The shortage of human resources in Video Surveillance creates severe challenges for monitoring the vast amounts of surveillance cameras present in the public domain. Therefore, to maintain public order and safety, security agents often rely on AI assistance. However, current AI-based surveillance systems still apply on core computer vision architectures such as Convolutional Networks or Vision Transformers for Action Recognition or Detection. These baselines require extensive finetuning to achieve good results in downstream tasks. This problem is further exacerbated in Surveillance, where the large range of scenarios, difficult viewing angles, low image quality and long format of videos, combined with the difficulty for setting up new datasets, make extensive finetuning extremely difficult \cite{BEAR}. Consequently, it is imperative to devise flexible approaches that can easily extend to different scenarios in zero or few-shot settings.

The goal of this work is to leverage the strong zero or few-shot generalization performance of Large Language Models (LLMs) in down-stream tasks~\cite{llama} and evaluate their potential for Video Understanding in Surveillance. We explore Large Vision-Language Models (LVLMs)~\cite{VideoLLaMA2}, which are capable of simultaneously processing vision and language by extending LLMs with a vision encoder, projecting its outputs to a common token embedding space~\cite{LLaVA}. Despite significant progress in the field, LVLMs still struggle with reasoning over videos, especially in the long format, due to their added temporal complexity, which is difficult to model in the shorter context size windows of smaller and mid-sized LLMs. To handle this, LVLMs typically sample frames linearly, in either high or low volume. 

\begin{figure}[t]
    \centering
    \includegraphics[width=\linewidth]{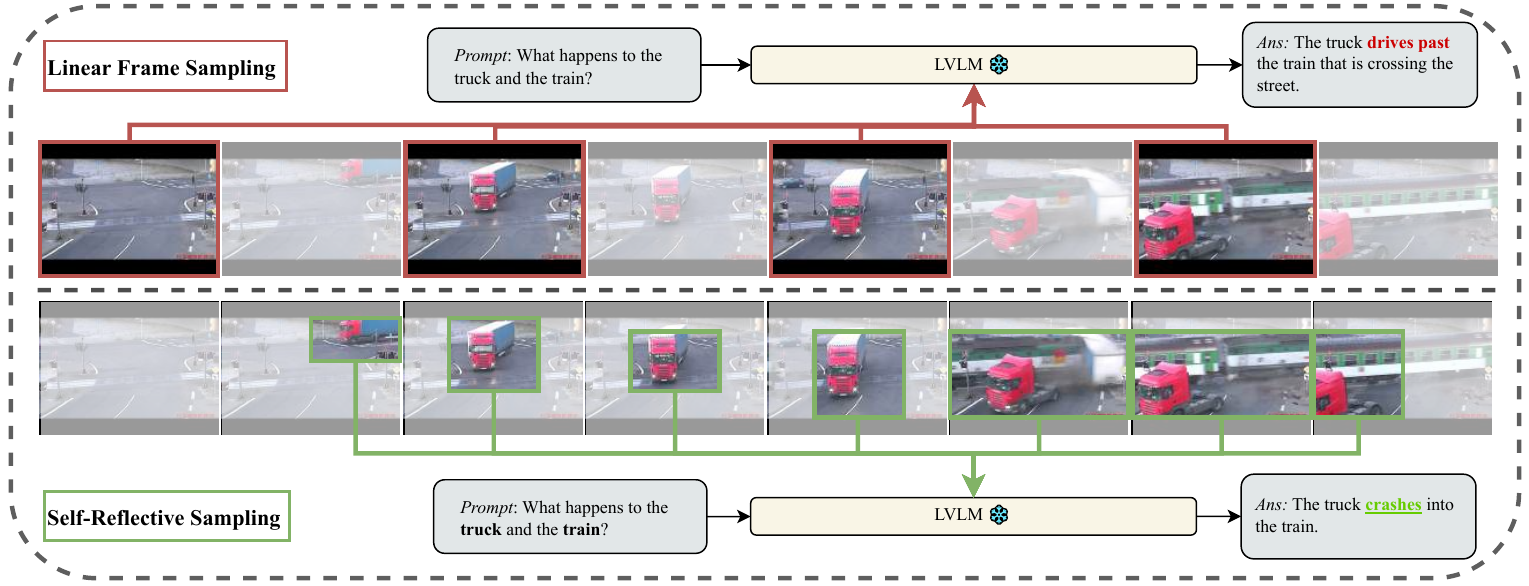}
    \caption{\textbf{Linear Frame Sampling vs Self-Reflective Sampling.} Adapted from \cite{Self-ReS}. In surveillance, events often occur in a non-linear timeline, and occupy a small spatial-temporal region of the video.}
    \label{fig:teaser}
\end{figure}

We argue that token-level selection is a superior approach for surveillance use cases, where capturing relevant events requires a combination of 1) fine-grained spatial and temporal localization, and 2) dealing with nonlinear events over long videos, as illustrated in Figure~\ref{fig:teaser}.

To test our hypothesis, we conduct experiments using a state-of-the-art LVLM - VideoLLaMA2~\cite{VideoLLaMA2} - in a widely recognized surveillance dataset - UCF Crime~\cite{ucfcrime}- and report results on two different sampling strategies: the default linear frame sampling setting, where the model samples a small, fixed number of frames; and Self-Reflective Sampling, a conditional self-reflective token selection strategy.
\renewcommand{\thefootnote}{}
\footnotetext{This work is supported by NOVA LINCS ref. UIDB/04516/2020 (https://doi.org/10.54499/UIDB/04516/2020) and ref. UIDP/04516/2020 (https://doi.org/10.54499/UIDP/04516/2020) with the financial support of FCT.IP; and Fundação para a Ciência e Tecnologia ref. 2023.03647.BDANA}
\renewcommand{\thefootnote}{\arabic{footnote}}
\section{Methodology}

\paragraph{Large Vision and Language Models.} We adopt the framework of Large Vision and Language models due to their strong performance in downstream tasks~\cite{VideoMME}. These models typically implement variations of the architecture in LLaVA~\cite{LLaVA}, which consists of: 1) a pretrained vision-language encoder such as CLIP~\cite{CLIP}, used to extract language-aligned visual features from the input image or video; 2) a linear projection matrix that linearly projects these visual tokens into the same embedding space as the LLM's textual input; and 3) a decoder-only Large Language Model, which reasons over input visual and textual features and autoregressively predicts an answer to a user query regarding the visual input.

\begin{figure}[t]
    \centering
    \includegraphics[width=\linewidth]{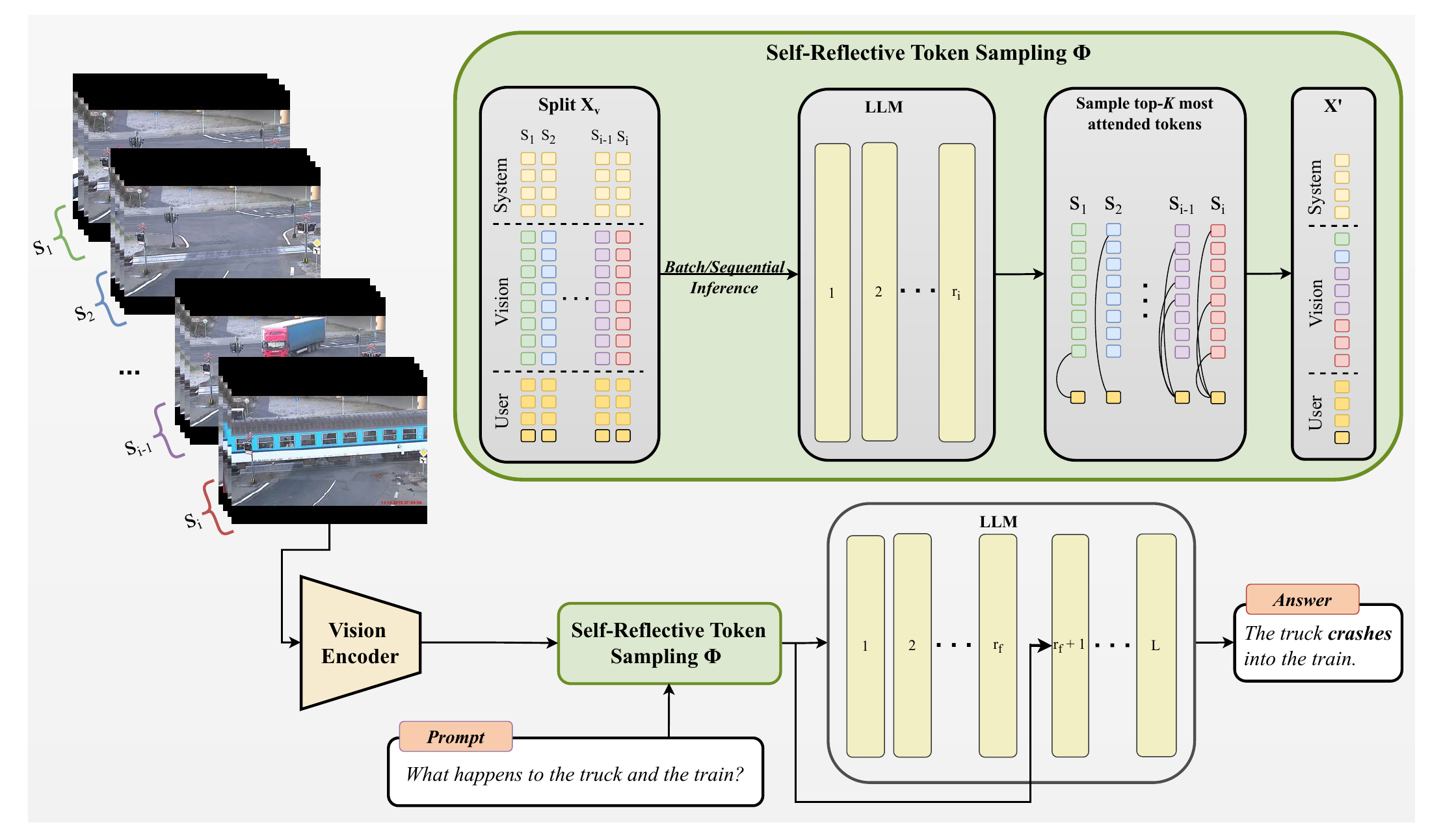}
    \caption{\textbf{Self-Reflective Sampling.} Adapted from \cite{Self-ReS}. Self-ReS allows the LLM to self-reflectively sample the top most attended tokens by the LLM.}
    \label{fig:selfres}
\end{figure}

\paragraph{Sampling Methods} We experiment with two different sampling methods of different granularity (Figure \ref{fig:teaser}). For a baseline measurement, we use the default LVLM sampling method of selecting a small set of frames based on a linear pattern (e.g. uniformly distributed). We also integrate a token-level Self-Reflective Sampling~\cite{Self-ReS} method (Figure \ref{fig:selfres}) that samples the most relevant tokens for the user query. Self-ReS begins by increasing the volume of sampled frames and splitting the video into segments with the same size as the model's default frame count. These segments then follow the normal procedure and are processed by the vision encoder to obtain visual features, and linearly projected. Self-ReS operates at the level of the LLM, by processing input features $\mathbf{X} \in \mathbb{R}^{N_s \times N' \times d}$, where $N_s$ is the number of segments, $N'$ is the aggregated number of system, visual, and user tokens and $d$ is the size of the LLM's hidden states:
\begin{equation}
    \mathbf{X} = \{[X_{sys}, X_{v}^{s_i}, X_{user}]\}_{i=1}^{N_s}. 
\end{equation} 
$X$ is forwarded through a set of self-reflective LLM layers $R = \{r_1, \ldots, r_j\}$, where each layer $r_j$ performs a \textit{Self-Reflective Sampling} step, denoted as $\Phi$. At each layer $r_j$, a pruning factor $p_j \in P$ controls the number of visual tokens retained, with $P=\{p_1, \ldots, p_j\}$. The sampling step $\Phi$ selects the top $p_j \times N$ visual tokens with the highest similarity to the final input token, which primarily attends to a sparse subset of relevant visual tokens \cite{FastV}. This effectively prunes the segments from $N_s$ to $N_{S'} = \frac{N_s}{p_j}$. Self-ReS iteratively updates the set of visual tokens by propagating them through the self-reflective layers $r_j \in R$, progressively reducing the number of segments and retaining the most relevant visual tokens. The final output is a spatio-temporal signature of the input video, conditioned on the user's query.

\section{Results}

\paragraph{Experimental Setup} We select VideoLLaMA2~\cite{VideoLLaMA2} as the baseline LVLM for our experiments (namely the 7B model), due to being one of the best performing models in video and language tasks, and because it has been shown to work well with Self-ReS~\cite{Self-ReS}. We adopt linear frame sampling using VideoLLaMA2's default 16 frames, and also consider different parameters for Self-ReS (Table \ref{tab:accuracy}). Experiments are conducted in a widely recognized surveillance dataset - UCF-Crime~\cite{ucfcrime} - in the task of Action Recognition for a subset of the original 13 classes of UCF-Crime: Abuse, Arrest, Arson, Assault, Burglary, Fighting, Robbery, Shooting, Stealing, Shoplifting, and Vandalism. Remaining classes are omitted as the goal is to measure the capability of the model in human related incidents only, for direct comparison with other reported zero-shot vision-language models (Table \ref{tab:accuracy}).  

\begin{figure}[t]
    \centering
    \includegraphics[width=\linewidth]{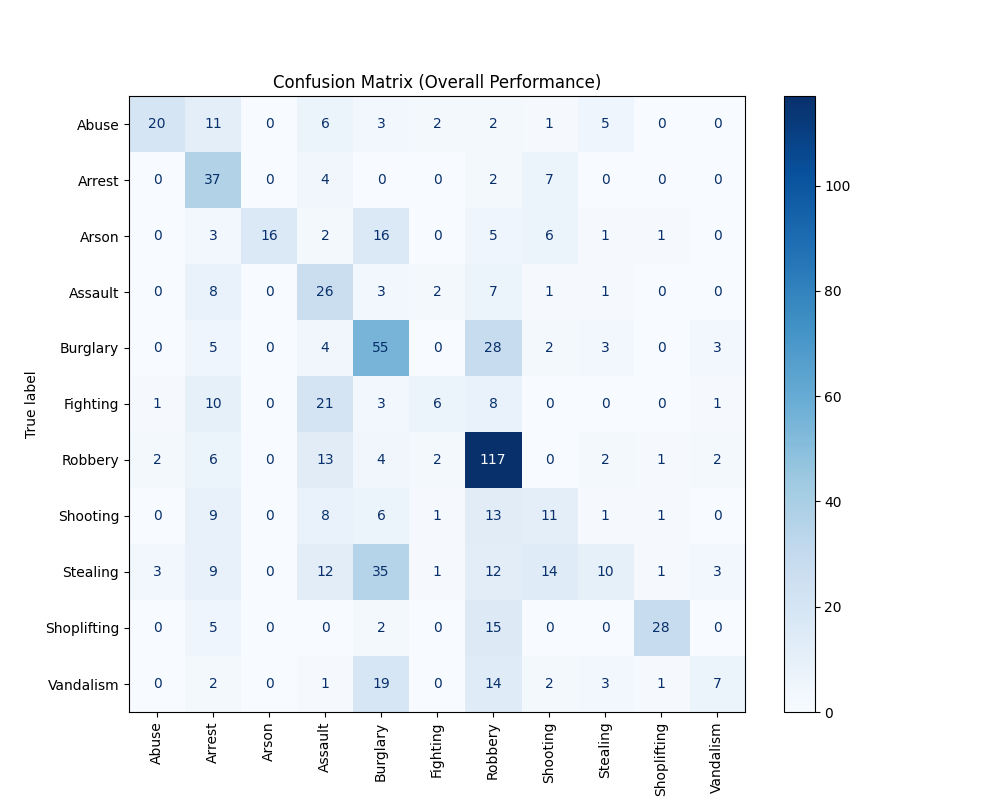}
    \caption{\textbf{Confusion Matrix of VideoLLaMA2~\cite{VideoLLaMA2} results in UCF-Crime~\cite{ucfcrime}.} The model performs well, but fails in situations where action definitions are very similar.}
    \label{fig:conf_Self}
\end{figure}

\paragraph{Discussion} Results can be seen in table \ref{tab:accuracy}. It is clear that Large Vision-Language Models represent a much stronger approach for zero-shot recognition in comparison to previous methods such as CLIP~\cite{CLIP}. This is natural, since LVLMs apply stronger and more recent encoder-only VL backbones in conjunction with an LLM. We estimate that this performance is still capped by the fact that image quality in UCF-Crime is very low, and LVLMs benefit from high resolution input \cite{LLaVA}. We verify that an improved sampling method can result in additional incremental improvement of $2\%$. The confusion matrix in Figure \ref{fig:conf_Self} provides further insights into the effectiveness of the language model, and the additional challenges that it introduces. In surveillance, different categories may be very similar in nature but differ in minor details. We can easily verify this in the confusion matrix \ref{fig:conf_Self}, where the model shows some difficulties in distinguishing between Burglary, Stealing and Robbery; as well as Fighting and Assault, where the difference is that in Assault, the victim does not fight back. We estimate that the similarities present in video content further extend to the language domain, and that the model may have been trained on data that uses these terms interchangeably.

\section{Future Work}
We strongly believe that Large Vision-Language Models are the future of surveillance, due to their zero-shot capabilities for diverse scenarios and their strong baseline performance. In the future, more mature solutions will be able to leverage in context learning and provide the LLM with examples that help it distinguish between the different categories. This will be advantageous for both recognizing abnormal situations and for surveillance related question-answering. Additionally, improved sampling techniques will allow the model to view relevant information and ignore background noise.

\begin{table}[t]
    \begin{center}
        {\small{
            \begin{tabular}{lccc}
                \toprule
                \textbf{Method} & $\mathbf{N_s}$ & $\mathbf{r_j}$ & \textbf{Accuracy} \\ 
                \midrule
                CLIP~\cite{CLIP} & - &-& 24.3\\ 
                VideoLLaMA2~\cite{VideoLLaMA2}  & 1 &  - & 42.6 \\ 
                \midrule
                \multirow{4}{*}{+ Self-ReS Regular} & 3 & 8 & \underline{42.8} \\
                 & 3 & 12 & 41.4 \\
                 & 5 & 8 & 42.6 \\
                 & 5 & 12 & 39.7 \\
                \midrule
               \multirow{3}{*}{+ Self-ReS Smooth} & 5 & 3 & 42.1 \\
                 & 5 &  5 & \textbf{44.6} \\
                 & 5 &  8 & 40.7 \\
                \bottomrule
            \end{tabular}}
        }
    \end{center}
    \caption{\textbf{Performance of VideoLLaMA2~\cite{VideoLLaMA2} in Action Recognition on UCF-Crime.} Results reveal that Large Vision-Language Models obtain much superior performance than CLIP~\cite{CLIP, BEAR} in a zero-shot setting (20\%+), establishing a new baseline. Self-ReS \cite{Self-ReS} results in a 2\% accuracy increase.} 
    \label{tab:accuracy}
\end{table}

\paragraph{}
\bibliography{recpad2k}
\end{document}